\documentclass{article}

\usepackage{microtype}
\usepackage{graphicx}
\usepackage{subfigure}
\usepackage{booktabs} 

\usepackage{hyperref}

\usepackage[accepted]{mlsys2025}

\mlsystitlerunning{FreeScale: Distributed Training for Sequence Recommendation Models with Minimal Scaling Cost}

\begin{document}

\twocolumn[
\mlsystitle{FreeScale: Distributed Training for Sequence Recommendation Models with Minimal Scaling Cost}



\mlsyssetsymbol{equal}{*}

\begin{mlsysauthorlist}
\mlsysauthor{Chenhao Feng}{equal,meta}
\mlsysauthor{Haoli Zhang}{equal,meta}
\mlsysauthor{Shakhzod Ali-Zade}{meta}
\mlsysauthor{Yanli Zhao}{meta}
\mlsysauthor{Liang Luo}{meta}
\mlsysauthor{Jennifer Cao}{meta}
\mlsysauthor{Lisen Deng}{meta}
\mlsysauthor{Siqiao Chen}{meta}
\mlsysauthor{Chenyu Zhao}{meta}
\mlsysauthor{Tristan Rice}{meta}
\mlsysauthor{Daniel Johnson}{meta}
\mlsysauthor{Min Si}{meta}
\mlsysauthor{Tiantu Xu}{meta}
\mlsysauthor{Yi Zhang}{meta}
\mlsysauthor{Siqi Yan}{meta}
\mlsysauthor{Chuanhao Zhuge}{meta}
\mlsysauthor{Min Ni}{meta}
\mlsysauthor{Bi Xue}{meta}
\mlsysauthor{Qunshu Zhang}{meta}
\mlsysauthor{Shen Li}{meta}
\end{mlsysauthorlist}

\mlsysaffiliation{meta}{Meta Inc}

\mlsyscorrespondingauthor{Shen Li}{cs.shenli@gmail.com}

\mlsyskeywords{Recommendation Systems, Distributed Training,
GPU Clusters, Collective Communication, Code Generation}

\vskip 0.3in
\begin{abstract}
Modern industrial Deep Learning Recommendation Models typically extract user preferences through the analysis of sequential interaction histories, subsequently generating predictions based on these derived interests. 
The inherent heterogeneity in data characteristics frequently result in substantial under-utilization of computational resources during large-scale training, primarily due to computational bubbles caused by severe stragglers and slow blocking communications. This paper introduces FreeScale, a solution designed to (1) mitigate the straggler problem through meticulously load balanced input samples (2) minimize the blocking communication by overlapping prioritized embedding communications with computations (3) resolve the GPU resource competition during computation and communication overlapping by communicating through SM-Free techniques.
Empirical evaluation demonstrates that FreeScale achieves up to 90.3\% reduction in computational bubbles when applied to real-world workloads running on 256 H100 GPUs.

\end{abstract}
]



\printAffiliationsAndNotice{\mlsysEqualContribution} 

\section{Introduction}
Deep Learning Recommendation Models (DLRMs)~\cite{DLRM, luo2024disaggregatedmultitowertopologyawaremodeling} have become the foundation for numerous industrial applications, including e-Commerce platforms, social media networks and digital advertising systems~\cite{hstupresentation, zeng2024accelerating, steck2021deep, zhao2021dear}. 
These models are designed to predict user preferences by analyzing interaction patterns within historical data. 
The efficacy of these models is largely contingent upon their ability to accurately extract user interests from extensive historical interaction sequences.

%
%
%
%
The volume and composition of Users' Interaction History (UIH) 
exhibit significant heterogeneity across the user population.~\cite{liu2023linrec, wang2021survey}. 
This inherent variability introduces computational stragglers and inefficient communication patterns during model training scaling up.
The stragglers emerge when workloads with substantially different execution times are allocated to different computational processes, causing processes with lighter workloads to remain idle while awaiting the completion of more computationally intensive ones. 
Meanwhile, the inefficienct communication patterns are due to blocking communication operations for embedding lookups that prevent concurrent computation, creating additional idle periods.
In large-scale training involving hundreds of GPUs, these inefficiencies leads to extended training durations and significant financial waste. 
Addressing these inefficiencies has become critical to maintaining a high return on investment.

This paper introduces FreeScale, a holistic solution designed to systematically minimize the scaling cost in general distributed deep learning based recommender system training, which consists of:
\begin{itemize}
\item\textbf{Sequence Load Balancing}: FreeScale balances input samples 
by estimating their computational complexity
to mitigate stragglers,
ensuring more uniform execution times across all processes.
\item\textbf{Prioritized Embedding Updates}: FreeScale identifies collision rows between consecutive training iterations to minimize blocking communications, allowing for greater overlapping with computation operations.
\item\textbf{SM-Free Communication}: FreeScale communicates via CPU-RDMA, resolving GPU resource competition issue during communication and computation overlapping.
\end{itemize}

\begin{figure*}[th!]
    \centering
    \includegraphics[width=\linewidth]{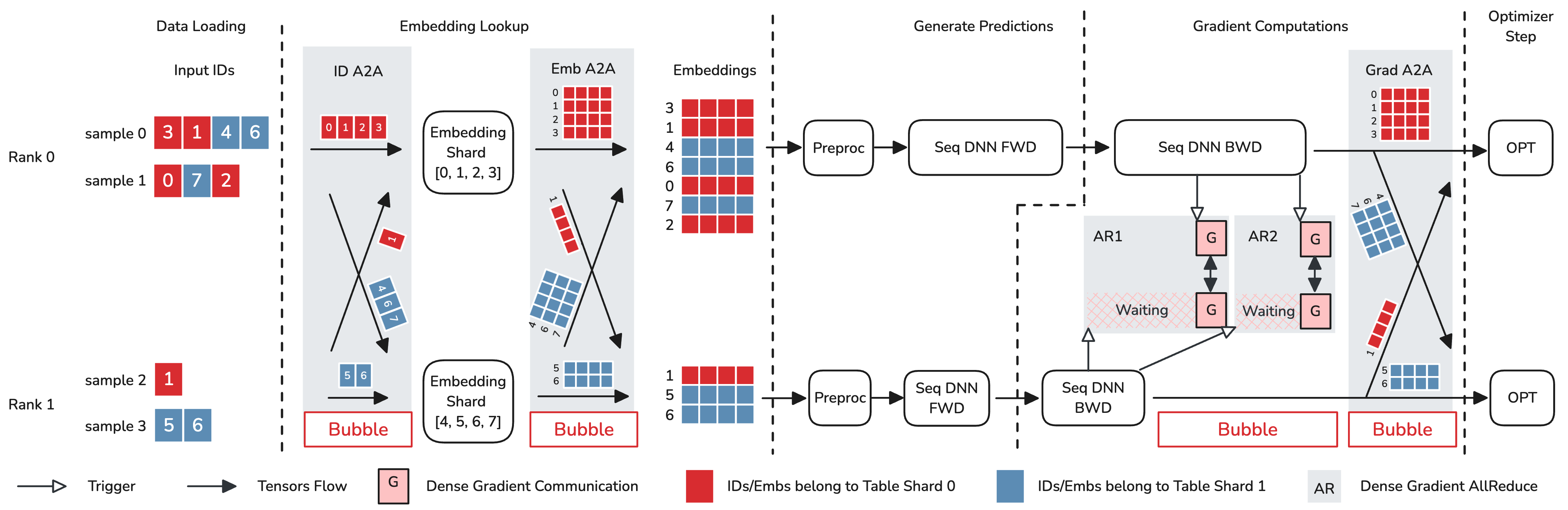}
    \caption{Simplified Sequence Model Training Iteration}
    \label{fig:background_dlrm}
\end{figure*}

FreeScale is implemented on top of PyTorch~\cite{pytorch:nips:2019} and TorchRec~\cite{torchrec}, with customized Triton~\cite{tillet2019triton} kernels. 
Evaluation on real-world dataset with a cluster of 256 NVIDIA H100 GPUs demonstrates that FreeScale reduced exposed communications by 90.3\% compared to vanilla TorchRec.

The remainder of this paper is organized as follows. 
Section~\ref{sec:background} presents the background of efficiency challenges in recommendation systems. 
Section~\ref{sec:design} details the underlying components of FreeScale. 
Implementations are discussed in Section~\ref{sec:implementation}. 
Section~\ref{sec:eval} presents our experimental setup and performance evaluation results. 
Section~\ref{sec:discussion} illustrates more of FreeScale's engineering analysis and SM Free communication.
Section~\ref{sec:related} summarizes related work, and Section~\ref{sec:conclusion} concludes the paper.

\section{Background}\label{sec:background}

This section briefly introduces sequence modeling in DLRMs, and highlights two unique efficiency concerns in distributed training for sequence models.

\subsection{Sequence Modeling in Recommendation Systems}\label{sec:seq_modeling}
Depending on the problem, sequence has different meanings. 
In factorization-machine based recommendation \cite{zhang2024wukong, luo2024disaggregatedmultitowertopologyawaremodeling, DHEN}, sequences represent abstract content IDs of past user interactions. 
The ID list is converted to a sequence of embeddings and a summarization architecture is applied to learn hidden information between user-item interations \cite{zeng2025interformereffectiveheterogeneousinteraction,AI_at_Meta_2023}.
In generative recommendation, sequences serve as the flattened token list and the model outputs for the action tokens of whether the user interacted with the current item for the ID tokens, transducing the recommendation task into a next-token prediction task \cite{zhai2024actionsspeaklouderwords}. 
Alternatively, as shown in \cite{wang2024generativerecommendationnextgenerationrecommender}, the sequence can represent organic content as well (e.g., a tokenized text post). 

Since our proposed techniques apply to all these settings, in this paper, we do not our limit ourselves to specific problem formulations.

To provide a concrete example of sequence model training, 
Figure~\ref{fig:background_dlrm} illustrates a single distributed training iteration. 
Each computational process exclusively operates on a dedicated GPU, denoted by a unique rank.
This example reflects how a typical model parameters distribution works in the recommendation systems,
which consists of massive sparse embedding tables alongside substantially smaller dense Deep Neural Network (DNN) layers. 
Under such settings, the embedding tables are partitioned while the DNN layers are replicated across GPUs.
Each rank independently retrieves inputs with UIH sequences of embedding IDs. During forward, these IDs are distributed to their corresponding partition through an \texttt{AllToAll} operation to perform embedding lookup, after which the resulting values are returned to their originating rank via a second \texttt{AllToAll} operation.
During backward, the dense gradients are synchronized using \texttt{AllReduce} operations \cite{ddp}, while embedding gradients are shuffled to their respective partitions through \texttt{AllToAll}.

Due to the nature of collective communications, the straggler rank forces other ranks to wait.
In this example, rank 0 processes more IDs and hence becomes the straggler of this iteration, forcing rank 1 to wait for rank 0's operations.
Also, blocking \texttt{AllToAll} communication on the critical path forces subsequent operations to wait. 
Scaling out to larger clusters will exacerbate these 2 issues.

\subsection{Imbalanced Workload and Stragglers}\label{sec:background_straggler}

\begin{figure}
\begin{minipage}[c]{\linewidth}
\centering
    \begin{minipage}[c]{0.49\linewidth}
    \includegraphics[width=\linewidth]{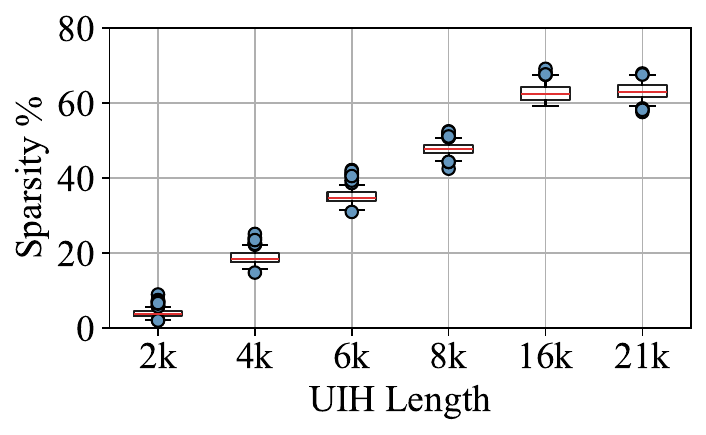}
    \end{minipage}
    \begin{minipage}[c]{0.49\linewidth}
    \includegraphics[width=\linewidth]{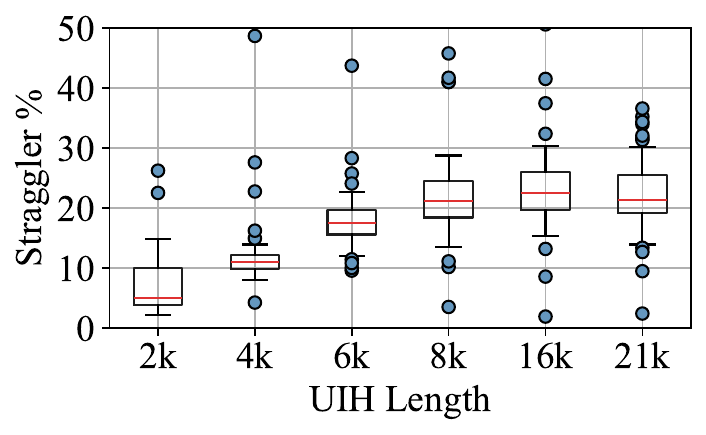}
    \end{minipage}
\end{minipage}
\begin{minipage}[c]{\linewidth}
\centering
    \begin{minipage}[c]{0.49\linewidth}
    \includegraphics[width=\linewidth]{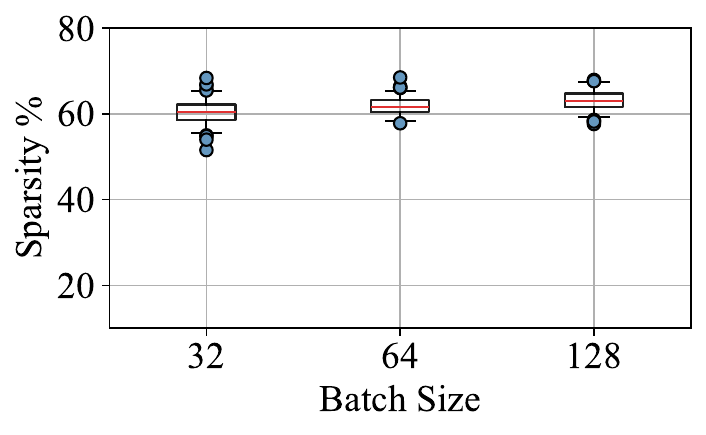}
    \end{minipage}
    \begin{minipage}[c]{0.49\linewidth}
    \includegraphics[width=\linewidth]{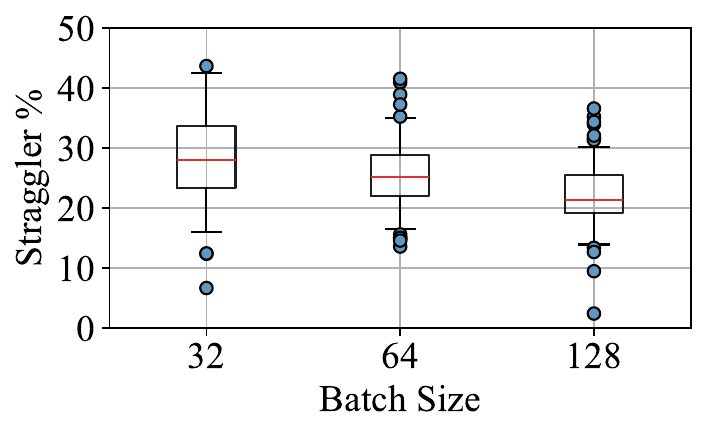}
    \end{minipage}
\end{minipage}
\begin{minipage}[c]{\linewidth}
\centering
    \begin{minipage}[c]{0.49\linewidth}
    \includegraphics[width=\linewidth]{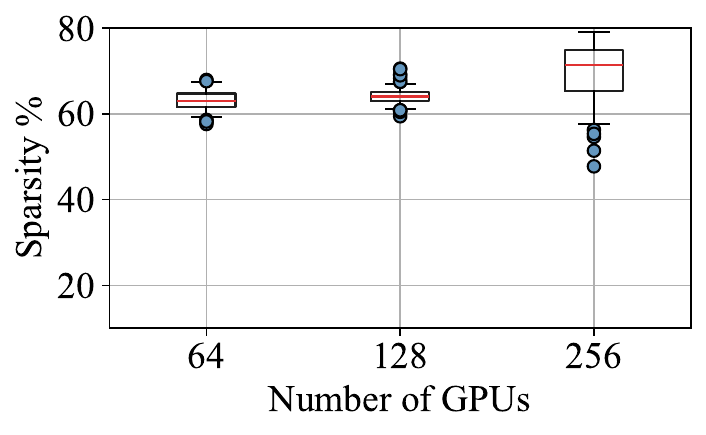}
    \end{minipage}
    \begin{minipage}[c]{0.49\linewidth}
    \includegraphics[width=\linewidth]{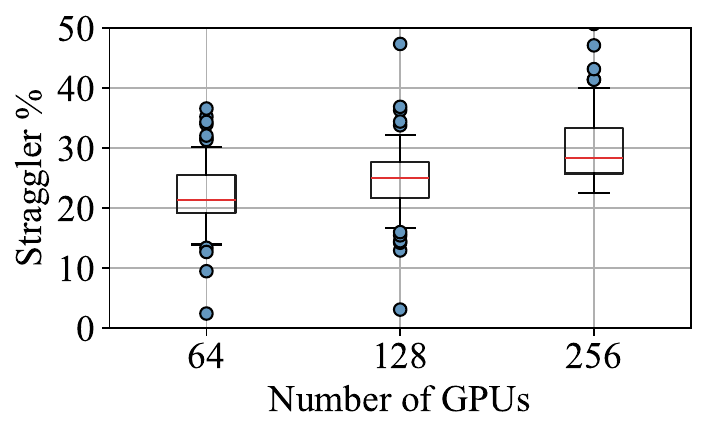}
    \end{minipage}
\end{minipage}
    \caption{Sparsity and Straggler Percentage}
    \label{fig:background_straggler}
    \vspace{-1.5em}
\end{figure}

Figure~\ref{fig:background_straggler} presents empirical measurements of local batch sparsity and global straggler effects in DLRM training on real-world production workload. 
Sparsity, quantified according to Equation~\ref{eq:sparsity}, is calculated as the proportion of padded items required when normalizing all samples to the maximum UIH length within a given batch. 
The straggler percentage metric represents the ratio of the mean idle wait time across all ranks relative to the total duration of the iteration. 
Unless otherwise specified, all experiments employ 21,000 max UIH length, 128 batch size, and 64 H100 GPUs. 

\begin{equation}
sparsity = 1 - \frac{\sum_{i=1}^{B} |uih_{i}|}{\max(|uih_{i}|, \forall i \in [1, B]) \times B}
\label{eq:sparsity}
\end{equation}

The observed sparsity monotonically increases with the UIH length up to approximately 16,000 items, beyond which the metric stabilizes, as few samples come with exceptionally long interaction histories. 
Sparsity demonstrates modest increases with both the batch size and the cluster scale, suggesting that the distribution of UIH lengths maintains relative consistent distributions. 
The straggler-induced resource under utilization exceeds 20\% when the maximum permitted UIH length surpasses 8,000 elements, regardless of variations in batch size and cluster scale parameters.

There're conventional approaches such as dense padding or sequence truncation to overcome stragglers. 
However, the former forces unnecessary computational overhead and the latter sacrifices potentially valuable behavioral data from highly engaged users.
While large language model training utilises methodologies such as stratifying corpus elements by length into discrete training phases and implementing sequence or context parallel across computational nodes, these approaches are not applicably in recommendation contexts. 
The temporal ordering of training samples in recommendation systems significantly influence model quality, thereby prohibiting the solution of length-based sample clustering. 
Furthermore, DLRMs are architecturally optimized for high-throughput processing of substantial traffic volumes, which usually contains relatively light computations where the additional communication overhead introduced by sequence or context parallelism would outweigh potential benefits.
Evidently, DLRMs require innovations to efficiently manage workload imbalances arising from variable sequence lengths.

\vspace{-1em}

\subsection{Embedding Communications and Row Collisions}\label{sec:background_collision}

\begin{figure}
\begin{minipage}[c]{\linewidth}
\centering
    \begin{minipage}[c]{0.49\linewidth}
    \centering
    \includegraphics[width=\linewidth]{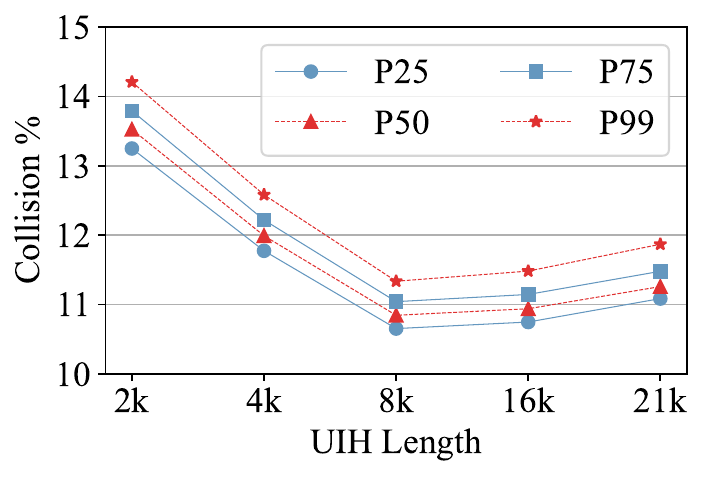}
    \end{minipage}
    \begin{minipage}[c]{0.49\linewidth}
    \centering
    \includegraphics[width=\linewidth]{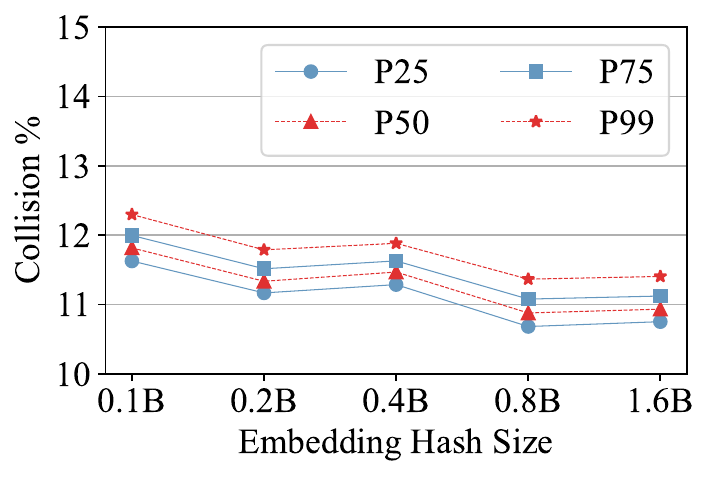}
    \end{minipage}
\end{minipage}
    \caption{ID Collision Percentage}
    \label{fig:background_collision}
    \vspace{-1em}
\end{figure}

Blocking embedding communications force subsequent computational operations to wait, leaving the computational resources under-utilized. 
A straightforward solution is to prefetch IDs and overlap the embedding lookup of the next iteration with the current computations.
However, such design leads to the next iteration consuming stale embedding values, as embedding lookup takes place before embedding row updates. 
Although the ratio of collisions is relatively low, 
it will potentially result in unpredictable degradations. 
In production, even marginal regressions of 0.1\% can translate to huge revenue losses, which is unacceptable.
Figure~\ref{fig:background_collision} visualizes the collision percentages derived from real production data,
indicating that collision percentages consistently maintain relatively modest levels across all examined configurations. 
This suggests that an ideal solution would prioritize embedding updates, which only waits for collision rows while proactively prefetching all non-collision rows in advance.

\vspace{-1em}
\section{Design}\label{sec:design}

FreeScale minimizes the scaling cost by avoiding computational stragglers and prioritizing embedding updates and communicates via SM-Free techniques to avoid occupying SMs for communications overlapped with computation operations. 

\subsection{Sequence Model Load Balancing}\label{sec:design_lb}

We propose to balance model dense layer execution times across all ranks by redistributing input samples, which would in turn minimize the size of bubbles on computational resources caused by dense gradient synchronizations.

\begin{figure*}[!htb]
    \centering
    \includegraphics[width=0.9\linewidth]{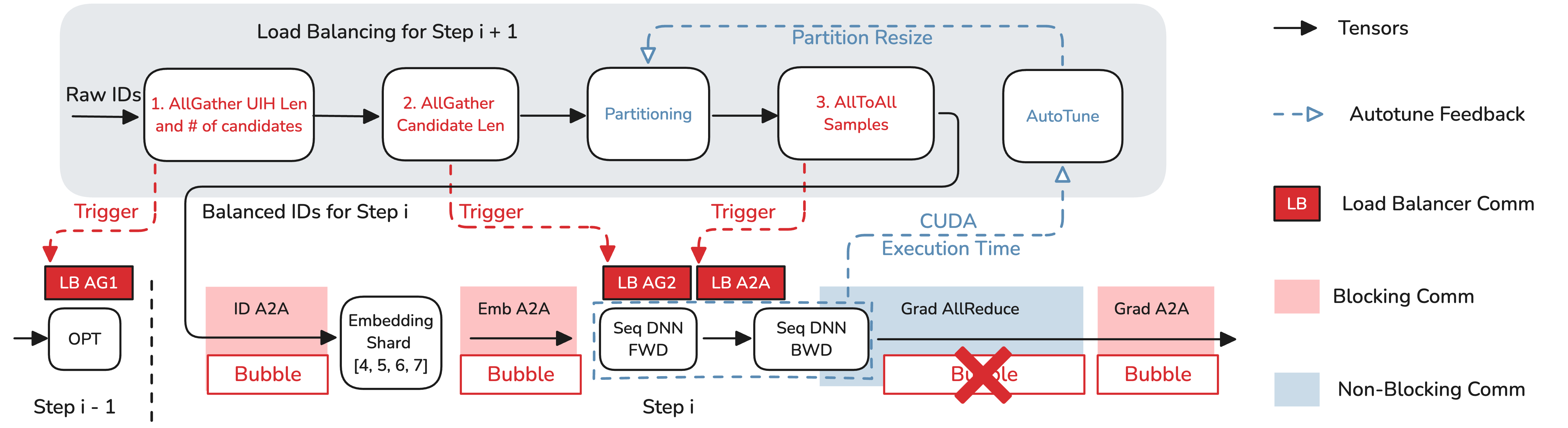}
    \caption{Load Balancing Example}
    \label{fig:lb_design}
    \vspace{-1em}
\end{figure*}

A naive solution would explore the feasibility of structuring data storage to achieve inherent balance prior to consumption by training processes. 
Unfortunately, this approach is impractical due to multiple constraints in production.
First, 
a specific model may ingest data from heterogeneous sources, while simultaneously, data from a single source may serve multiple contexts. 
To minimize storage overhead, training process typically retrieve raw data from multiple sources and dynamically transform them into appropriate formats, rather than maintaining redundant data copies.
Second, training processes are frequently deployed within dynamic resource allocation frameworks, wherein both the quantity and the type of GPUs vary according to incoming computational demands and the priority of the job. 
The dynamics in computational resources preclude static data partitioning strategies.
Third, optimal load balancing decisions often require calibration based on actual model execution metrics, which are inherently unavailable until runtime. 
Therefore, effective load balancing mechanisms should be integrated proximally to the model training procedure.

Incorporating load balancing algorithms directly within the training process requires a judicious design that satisfies at minimum two requirements:
\begin{itemize}
\item \textbf{High Efficiency}: The load balancer must operate with minimal costs and avoid disrupting or slowing down any other trainer components, as even marginal overhead on GPU resources can result in significant operational cost at scale. 
\item \textbf{Non-Intrusive Integration}: Infrastructure components and models are typically developed by different teams, with the latter requiring accelerated development cycles and frequent iterations. For broad adoption, the load balancer must impose minimal constraints on model behavior and decouple API invocations from model implementations.  
\end{itemize}
While standard DLRMs typically employ computationally intensive attention mechanisms and multi-layer perceptrons, the ID manipulation operations performed by the load balancer introduce relatively modest computational overhead, when properly implemented. 
The primary performance constraint stems from the inter-rank communication of IDs through collective operations. 
As discussed above, samples exhibit heterogeneity in multiple dimensions: variable-length UIH, variable candidate quantities, and variable content sizes per candidate. 
This heterogeneity forces a sequential communication pattern wherein each communication phase depends on the output of its predecessor, as outlined in Algorithm~\ref{algo:lb}. 
The load balancing mechanism initially aggregates the UIH lengths and the number of candidates from all ranks. This information facilitates the allocation of appropriately sized tensors to accommodate subsequent \texttt{AllGather}-ed world candidate lengths. Then, the system employs a configurable, deterministic partitioning algorithm \texttt{$\mathcal{P}$} that computes the balanced redistribution of samples. Finally, based on the partition result, all ranks join an \texttt{AllToAll} communication to shuffle samples to match the partitioning decision.
\begin{algorithm}
\caption{Load Balancing}\label{algo:lb}
\begin{algorithmic}[1]
\STATE Local input batch \texttt{$\mathcal{B}$} 
\STATE Number of trainer processes {WS}
\STATE Runtime execution time info {$R$}
\STATE Partitioning function \texttt{$\mathcal{P}$}
\STATE \textbf{Ensure} balanced output batch \texttt{$\mathcal{BB}$}
\STATE {$L_{uih}$ $\leftarrow$ Tensor([\emph{WS}, $\mathcal{B}$.size])} {\textcolor{blue}{\small{// world UIH lens}}}
\STATE {$N_{can}$ $\leftarrow$ Tensor([\emph{WS}, $\mathcal{B}$.size])} {\textcolor{blue}{\small{// world number candidates}}}
\STATE \textcolor{blue}{\small{Stage I: gather lengths}}
\STATE {AllGather($L_{uih}$, $\mathcal{B}$.uih\_lens)}
\STATE {AllGather($N_{can}$, $\mathcal{B}$.num\_candidates)}
\STATE {$L_{can}$ $\leftarrow$ Tensor([sum($N_{can}$)])} {\textcolor{blue}{\small{// world candidate lens}}}
\STATE \textcolor{blue}{\small{Stage II: gather composite candidate lengths}}
\STATE {AllGather($L_{can}$, $\mathcal{B}$.candidate\_lens)}
\STATE {send\_samples, recv\_samples = $\mathcal{P}$({$\mathcal{B}$}, {$L_{uih}$}, {$N_{can}$}, {$L_{can}$}, {$R$})}
\STATE \textcolor{blue}{\small{Stage III: shuffle samples}}
\STATE {AllToAll(send\_samples, recv\_samples)}
\STATE {$\mathcal{BB} \leftarrow \mathcal{B}$.from\_samples(recv\_samples)}
\STATE \textbf{Return} {$\mathcal{BB}$}
\end{algorithmic}
\end{algorithm}
Despite the sequential dependency across the three communication stages, it remains feasible to hide the communication latency by overlapping with the computation of DLRMs. This optimization is achievable because the training process typically maintains a prefetched buffer containing input batches for subsequent iterations.

Figure~\ref{fig:lb_design} provides a visual representation of this approach, with the three-phase communication sequences denoted by solid red rectangles. Through this design, the load balancing mechanism can effectively redistribute computational workloads across processing nodes without exposing the underlying communication overhead to the critical path. Consequently, this approach substantially mitigates computational bubbles in the computational resource that would otherwise result from imbalanced workloads. To fulfill the non-intrusive integration requirement, the load balancer implements its three-stage protocol as hooks, triggered at critical junctures within the training pipeline: during optimizer step execution, forward propagation, and backward propagation computations. These injection points are visually represented in Figure~\ref{fig:lb_design} by dashed red arrows. In popular machine learning frameworks such as PyTorch, these hooks can be easily installed into the execution graph via the established \texttt{nn.Module} APIs. Even in cases where such explicit hook interfaces are not directly exposed, the training pipeline orchestration typically falls under the purview of infrastructure teams, which are naturally decoupled from the modeling efforts. 

The design of the load balancer deliberately stays agnostic to specific partitioning algorithms, and therefore the discussions are postponed to the implementation sections. Later, Section~\ref{sec:partitioning} introduces two paradigms: fixed batch size partitioning and variable batch size partitioning, which provide sufficient coverage for most use cases we have encountered in real-world deployment scenarios.

\subsection{Prioritized Embedding Updates}\label{sec:design_cosu}

As analyzed in Section~\ref{sec:background_collision}, prefetching embedding values would theoretically eliminate these execution bubbles, but this approach introduces potential concerns regarding numerical stability and consequent regression of model quality. 
Based on this analysis, numerical parity with the synchronized baseline can be maintained by enforcing a strict write-read ordering on the collision rows, where communications on all other non-collision (\emph{i.e.}, exclusive) rows can be performed in an asynchronous fashion. 
This section focuses on row-wise sharded embedding tables, omitting column-wise use cases, as the latter exhibits uniform collision patterns across all ranks, thus presenting fewer design challenges. 

\begin{figure*}[!htb]
    \centering
    \includegraphics[width=\linewidth]{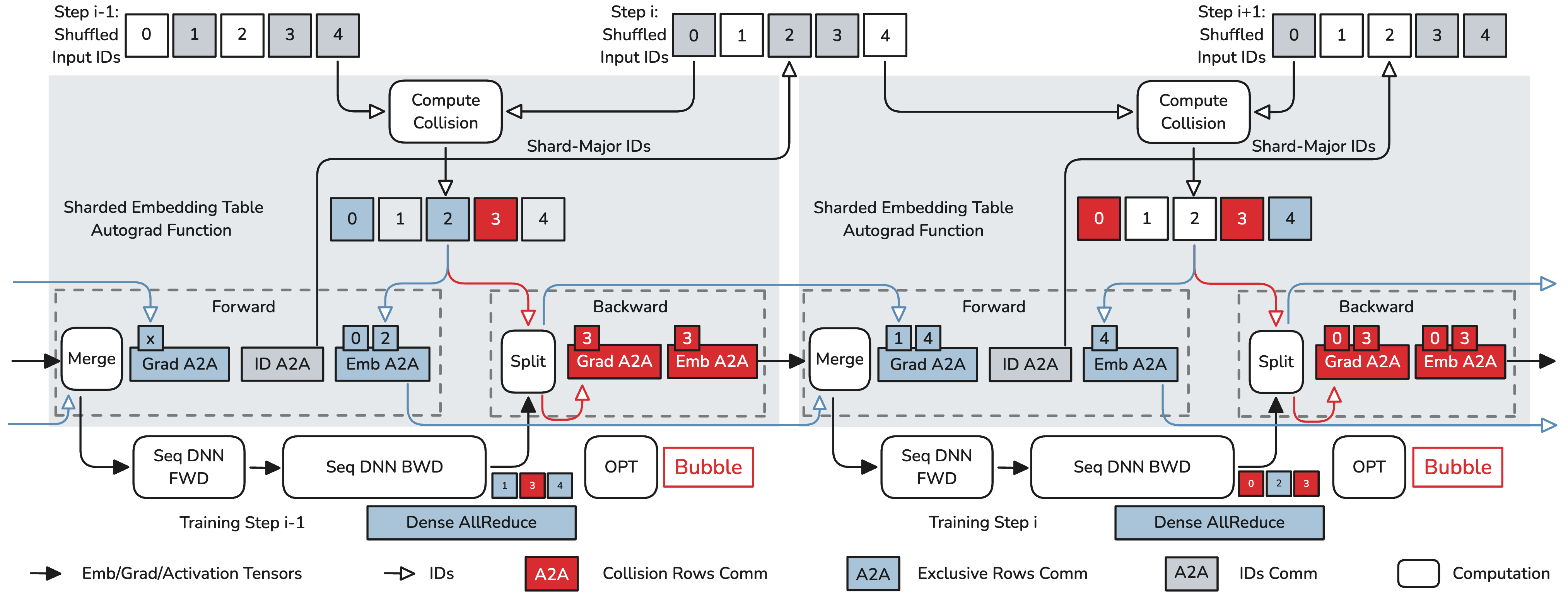}
    \caption{Prioritized Embedding Communication}
    \label{fig:cosu_design}
    \vspace{-1em}
\end{figure*}

Firstly, it's crutial to differentiate between collision rows and exclusive rows, which need to be calculated under the appropriate distribution form. 
In synchronized DLRM training, IDs manifest in two distinct structures: batch-major and shard-major. 
The input batch is initially structured in batch-major form, wherein IDs within each batch are processed by the same dense model replica, while their corresponding embeddings are distributed across different embedding shards. 
After the ID \texttt{AllToAll} communication phase, these IDs undergo transformation into a shard-major form, where all IDs on a particular rank belong to the same embedding shard, potentially originating from different ranks. 
Given that embedding values are stored in a shard-major structure, collision detection must likewise be computed in this corresponding form to attain correctness. 
This constraint prevents pre-computing collisions,
as the sharding strategy is determined at runtime based on available resources and configurations.

Since embedding tables are generally integrated within the model architecture as constituent components,
the logic governing prioritized embedding updates should be encapsulated within an embedding module, which subsequently replaces the original embedding table instance in the model object through module-swapping techniques~\cite{torchrec}. 
Under this paradigm, the algorithms can be carefully organized within the corresponding forward and backward methods of the \texttt{autograd.Function}~\cite{autograd} of the embedding module. 
This design ensures that the autograd engine will automatically trigger these methods with the appropriate timing when embeddings and their gradients become available for processing. 
The detailed approach is presented in Algorithm~\ref{algo:cosu}.

\begin{algorithm}[tb]
    \caption{Prioritized Embedding Autograd Function}
    \label{algo:cosu}
    \begin{algorithmic}[1]
        \STATE Persistent autograd context \textbf{$\mathcal{C}$} 
        \STATE Prefetched Embedding Indices, batch-major \texttt{$\tilde{I}^{i+1}$}
        \STATE Current Embedding Gradients, batch-major \texttt{$\tilde{G}^{i}$}
        \STATE \textbf{Forward}(\texttt{$\mathcal{C}$, $\tilde{I}^{i+1}$})
        \STATE \quad \textbf{with}  cuda.stream($\mathcal{C}$.stream)
        \STATE \qquad {$G_{ex}^{i-1} \leftarrow$ AllToAll($\mathcal{C}.\tilde{G}_{ex}^{i-1}$)} \textcolor{blue}{\small{// to shard-major}}
        \STATE \qquad {update\_embedding({$\mathcal{C}.I_{ex}^{i-1}$, $G_{ex}^{i-1}$})}
        \STATE \qquad {$\mathcal{C}.I^{i+1} \leftarrow$ AllToAll($\tilde{I}^{i+1}$)} \textcolor{blue}{\small{// to shard-major}}
        \STATE \qquad \textcolor{blue}{\small{// Compute collision and exclusive indices $I_{co}$, $I_{ex}$}}
        \STATE \qquad {$\mathcal{C}.I_{co}^{i}, I_{ex}^{i}, I_{ex}^{i+1} \leftarrow $ compute\_collision($\mathcal{C}.I^{i}$, $\mathcal{C}.I^{i+1}$)}
        \STATE \qquad \textcolor{blue}{\small{// Prepare exclusive embedding for step {$i+1$}}}
        \STATE \qquad {$\mathcal{C}.\tilde{E}_{ex}^{i+1} \leftarrow$ AllToAll(Lookup($I_{ex}^{i+1}$))} \textcolor{blue}{\small{// to batch-major}}
        \STATE \qquad {$\mathcal{C}.\tilde{I}_{co}^{i} \leftarrow $ AllToAll($\mathcal{C}.I_{co}^{i}$)
        \STATE \qquad $\mathcal{C}.\tilde{I}_{ex}^{i} \leftarrow $ AllToAll($I_{ex}^{i+1}$)}
        \STATE \quad \textbf{return} {merge($\mathcal{C}.\tilde{E}_{ex}^{i}$.wait(),$\mathcal{C}.\tilde{E}_{co}^{i}$.wait())}

        \STATE \textbf{Backward}({$\mathcal{C}$}, $\tilde{G}^{i}$)
        \STATE \quad \textbf{with} cuda.stream($\mathcal{C}$.stream)
        \STATE \qquad {$\tilde{G}_{co}^{i}$, $\mathcal{C}.\tilde{G}_{ex}^{i} \leftarrow $ split($\tilde{G}^{i}$, $\mathcal{C}.\tilde{I}_{co}^i$, $\mathcal{C}.\tilde{I}_{ex}^i$)}
        \STATE \qquad {$G_{co}^{i} \leftarrow $ AllToAll($\tilde{G}_{co}^{i}$)} \textcolor{blue}{\small{// to shard-major}}
        \STATE \qquad {$E_{co}^{i+1}\leftarrow$ update\_embedding($\mathcal{C}.I_{co}^{i}$, $G_{co}^{i}$)}
        \STATE \qquad {$\mathcal{C}.\tilde{E}_{co}^{i+1} \leftarrow $ AllToAll($E_{co}^{i+1}$)} \textcolor{blue}{\small{// to batch-major}}

    \end{algorithmic}
\end{algorithm}
For simplicity, Algorithm~\ref{algo:cosu} deliberately excludes \texttt{Tensor} size communications and focuses on index, embedding, and gradient communications, assuming that the \texttt{AllToAll} collective operation directly yields an appropriately dimensioned output \texttt{Tensor}. 
In our notation convention, tilde-adorned capital letters (\emph{i.e.}, indices \texttt{$\tilde{I}$}, embeddings \texttt{$\tilde{E}$}, gradients \texttt{$\tilde{G}$}) denote \texttt{Tensor}s in batch-major form, whereas their non-tilde counterparts represent the corresponding \texttt{Tensor}s in shard-major form. 
The superscripts adjacent to capital letters indicate the iteration number, while the subscripts specify whether the \texttt{Tensor} contains collision or exclusive values. 
As an example, \texttt{$G^{i}_{co}$} represents embedding gradients from iteration $i$ that collides with embeddings used in iteration $i+1$. 
The autograd context \texttt{$\mathcal{C}$} persists across all iterations, which allows one iteration to carry values across iteration boundaries\footnote{Garbage collection in the context is omitted for simplicity purposes}. 

The \texttt{forward} function accepts the persistent autograd context and the prefetched indices designated for the subsequent iteration. 
Initially, it completes the exclusive gradient synchronization from the preceding iteration and updates the corresponding rows accordingly. 
After that, it transforms the prefetched indices into shard-major format and computes collisions with the current iteration. 
The algorithm then retrieves exclusive embeddings and transforms them into batch-major format, preparing them for the next iteration. 
Finally, it converts indices from the current iteration into batch-major form to distinguish collision and exclusive gradients during the backward propagation phase. 
All aforementioned operations are queued into a dedicated \texttt{CUDA} stream to overlap with the model computations. 
The \texttt{forward} function awaits the completion of prefetched exclusive embeddings and collision embeddings from the previous iteration, merging these two components into one complete input batch to serve the current iteration.

The \texttt{backward} function starts with splitting the gradients of the current iteration into collision and exclusive components. 
Collision gradients are expedited to finish their update process, and the resulting updated embedding values immediately transform to a batch-major format to facilitate consumption by the next iteration's \texttt{forward} function. 
Figure~\ref{fig:cosu_design} illustrates this process across two consecutive iterations. 
Compared to the architecture depicted in Figure~\ref{fig:background_dlrm}, the exposed blocking \texttt{AllToAll} communication operation is substantially reduced, now encompassing only collision gradients and embeddings. 
This optimization should produce significant performance enhancements according to the empirical measurements presented in Section~\ref{sec:background_collision}.
\subsection{SM-Free Communication}

\begin{figure*}[!htb]
    \centering
    \includegraphics[width=\linewidth]{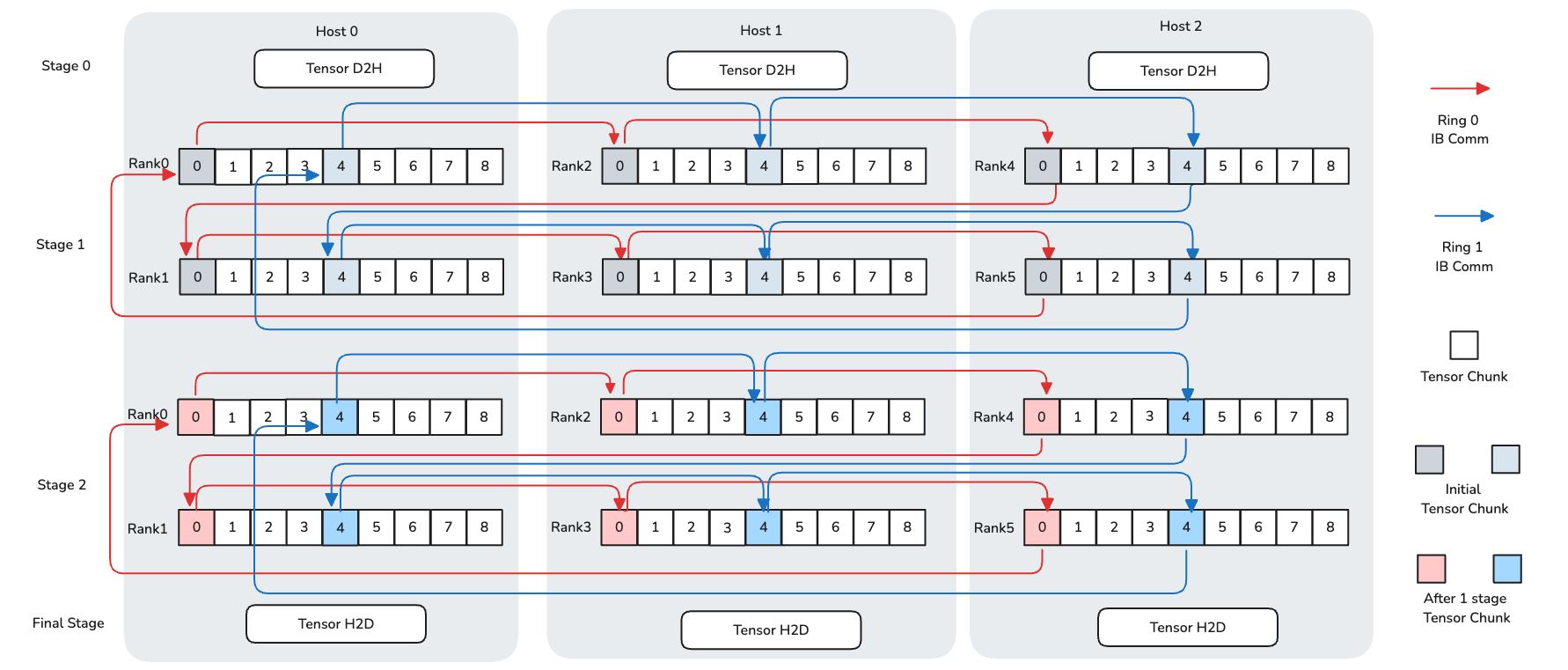}
    \vspace{-1em}
    \caption{SM-Free Communication}
    \label{fig:comm_design}
    \vspace{-1em}
\end{figure*}

Load balancing and prioritized embedding communication 
either introduces additional overlappable communications or transforms exposed communications into overlapped operations, 
where the communications are either \texttt{AllToAll} or \texttt{AllGather} without any value reductions. 
While these communications can be effectively concealed through computational overlapping techniques, the communication operations nevertheless consume Streaming Multiprocessor (SM) resources, 
thereby creating potential resource contention with computation kernels. 

To be noted that, even though NCCL uses GPU Direct RDMA that seems SM free, it's not autonomous ~\citep{hu2025demystifying}. 
NCCL still subdivides every collective into communication channels. Each channel is launched as a separate CUDA block that runs on its own SM.
Secondly, NCCL 2.28 introduced Copy Engine (CE) collectives to reduce SM utilization, but only within a single NVL domain ~\citep{nccl2283notes}. 
When the cluster size exceeds one NVL domain, NCCL falls back to implementation that uses SM. Additionally, it requires pre-registered memory (exposed via PyTorch's symmetric memory), which requires either careful memory planning or explicit buffer copies.

To circumvent occupying SMs, Figure~\ref{fig:comm_design} presents an \texttt{AllGather} communication example without SMs.
In this configuration, rank 0 establishes an RDMA ring topology with ranks across hosts. 
Initially, each rank transfers a chunk of its local data to host memory via D2H operations. 
Subsequently, it transmits this host memory chunk through RDMA operations to its neighbor on the ring following the clockwise direction.
In the subsequent stages, each rank propagates the chunk it received from the last stage on the RDMA ring to its clockwise neighbor.
This process continues iteratively until all chunks have been successfully received by all participating ranks. 
Finally, the tensor on host memory will be moved back via H2D operations.
In this way, non-reduction communications such as \texttt{AllGather} and \texttt{AllToAll} can be performed efficiently without consuming SM resources.

\section{Implementation}\label{sec:implementation}

This section presents the implementation details, beginning with two specific partitioning algorithms that underpin the load balancer, followed by Triton kernel optimizations for irregularly-shaped ID tensor manipulations. Finally, we explain how FreeScale integrated load balancing and prioritized embedding updates into the training pipeline while maintaining transparency for model engineers. 

\subsection{Partitioning Algorithms}\label{sec:partitioning}

FreeScale is compatible with arbitrary custom partitioning algorithms, as long as the algorithm maps global samples to ranks.  
We implemented two relatively generic approaches: Fixed Batch Size (FBS) and Variable Batch Size (VBS) partitioning methodologies. 
Both employ the UIH length as a computational complexity proxy to estimate execution time.
The FBS algorithm performs a global sorting of samples based on UIH length, subsequently assigning samples to distinct ranks through a zig-zag strategy, i.e., rank \texttt{i} gets samples with indices \texttt{[i, 2n-i, 2n+i, 4n-i, …]}, where \texttt{n} denotes the number of ranks in this distributed training job. 
This approach ensures that each rank receives an equivalent number of samples while simultaneously distributing samples with varying UIH lengths across ranks in a balanced manner.
The VBS algorithm adopts a more nuanced approach, calculating sample weights using the formula $L^{\alpha}$, where $L$ represents the UIH length and $\alpha$ as a hyperparameter. 
The algorithm subsequently partitions globally sorted samples into \texttt{n} segments, with each segment summing up to approximately equivalent total weight. 
Under the VBS approach, individual ranks may receive disparate sample quantities, while all samples within a given batch maintain similar lengths. 
Since the initial value of weight might not accurately represent the model computation execution time, VBS also employs an autotune mechanism that increments or decrements local batch size depending on the difference between the local execution time and the global average execution time, which is illustrated in Figure~\ref{fig:lb_design} as the \texttt{AutoTune} module. 
Each of these two partitioning strategy presents distinct trade-offs: 

\begin{itemize}
\item \textbf{FBS} offers versatility across models due to its independence from precise computational execution time estimation. 
However, when \texttt{CUDA} kernels use an individual block to cover one entire sample, FBS may inadvertently generate intra-kernel load imbalances. 
Blocks assigned to long UIH samples can emerge as intra-kernel stragglers that compel certain SMs to remain in an idle state while awaiting task completion. 
\item \textbf{VBS} constructs batches with samples of similar UIH lengths, significantly reducing intra-batch sparsity. 
This optimization enables kernel blocks to receive more uniformly distributed workloads, thereby enhancing overall computational efficiency. 
However, the implementation of dynamic batch size introduces additional complexities that preclude universal applicability. 
The modification of local batch size engenders subtle interactions with gradient synchronization mechanisms. 
Practitioners might need to calibrate loss function formulations and gradient collective communications (\emph{i.e.}, \texttt{AllReduce} and \texttt{AllToAll}) to ensure appropriate gradient scaling. 
\end{itemize}

These two built-in partitioning algorithms are sufficient to address most use cases encountered in our deployment experiences. 
For specialized scenarios requiring special workload distribution strategies, practitioners can plug in custom partitioning logic by supplying user-defined Python functions following the parameter specifications as shown in Algorithm~\ref{algo:lb}. 
This extensibility allows FreeScale to adapt to domain-specific workload characteristics without requiring modifications to the underlying distribution infrastructure.

\subsection{Triton Kernel Optimizations}

FBS and VBS partitioning strategies require advanced indexing and concatenation operations on irregularly shaped tensor structures~\cite{jagged}. 
Implementing these operations using standard PyTorch primitives leads to iterative processing of individual samples followed by subsequent concatenations, which is inefficient.

\begin{figure}
    \centering
    \includegraphics[width=\linewidth]{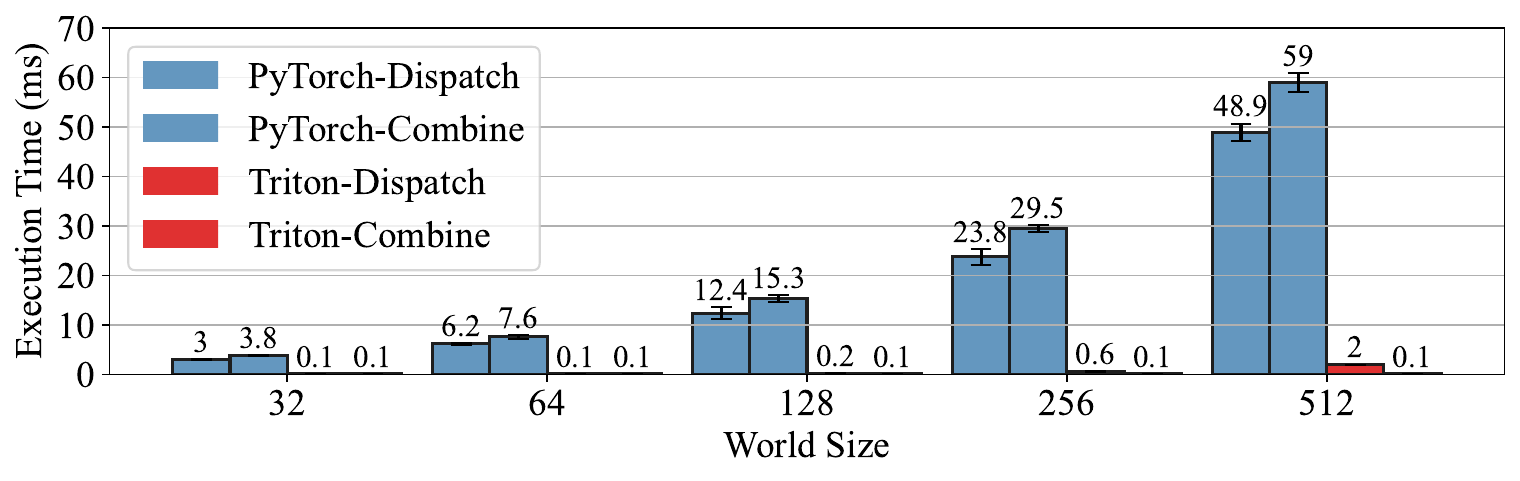}
    \vspace{-2em}
    \caption{Kernel Efficiency}
    \label{fig:triton}
    \vspace{-1.5em}
\end{figure}

To address these inefficiencies, we implemented a comprehensive suite of specialized Triton~\cite{tillet2019triton} kernels optimized for jagged tensor operations. 
These operations include the indexed permute kernel that apply partitioning transformations to globally \texttt{AllGather}ed IDs, the ranged dispatch kernel and the ranged combine kernel that prepare input and output tensors for ID and embedding \texttt{AllToAll} operations, and the keyed transpose kernel that converts jagged tensor representations from the feature-major form to the batch-major form, among others. 
Due to space constraints, we only present the performance metrics for the ranged dispatch and ranged combine kernels. 
Figure~\ref{fig:triton} illustrates a performance analysis between vanilla PyTorch implementations and our specialized Triton kernels. 
The execution latency of the PyTorch implementation increases linearly with the world size, becoming computationally prohibitive when the world size exceeds several hundred nodes. 
In contrast, the Triton implementation demonstrates superior performance, outperforming PyTorch eager implementations by a factor of 20 even at modest world sizes of 32, with efficiency differentials expanding to more than 600-fold at world size 512.

\subsection{Staged Training Pipeline}

The training framework usually faces an inherent dilemma. 
In one dimension, it must offer high flexibility to accommodate diverse model designs, which prevents the trainer from making specific assumptions of model architectures. 
Conversely, to achieve optimal computational efficiency, the training framework would benefit substantially from accessing detailed model architectural information.
The theoretical optimum would involve obtaining an operator graph by tracing the full train step, allowing the training framework to systematically manipulate this graph to implement performance-enhancing transformations. 
However, imposing full graph traceability introduces significant constraints on model development practices. 
Such constraints would prohibit computational components including dynamic branching logic and third-party tensor operators, thereby impeding development flexibility.

\begin{figure*}[!htb]
    \begin{minipage}[c]{0.33\textwidth}
    \centering
    \includegraphics[width=\linewidth]{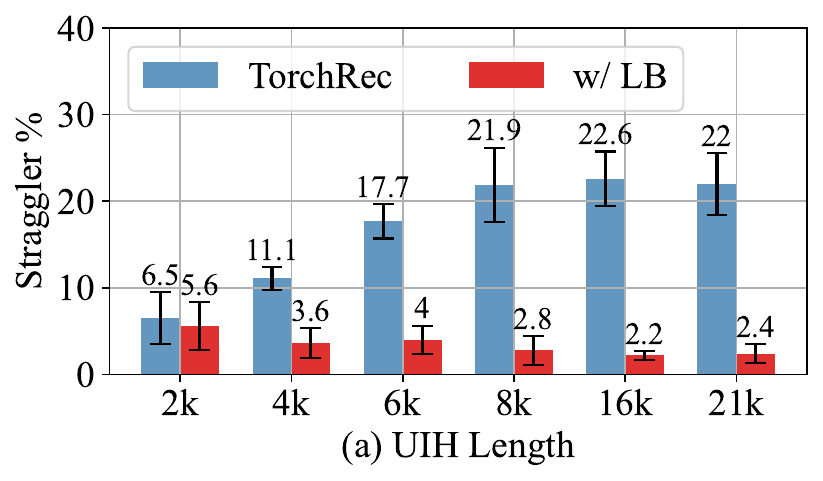}
    \end{minipage}
    \begin{minipage}[c]{0.33\textwidth}
    \centering
    \includegraphics[width=\linewidth]{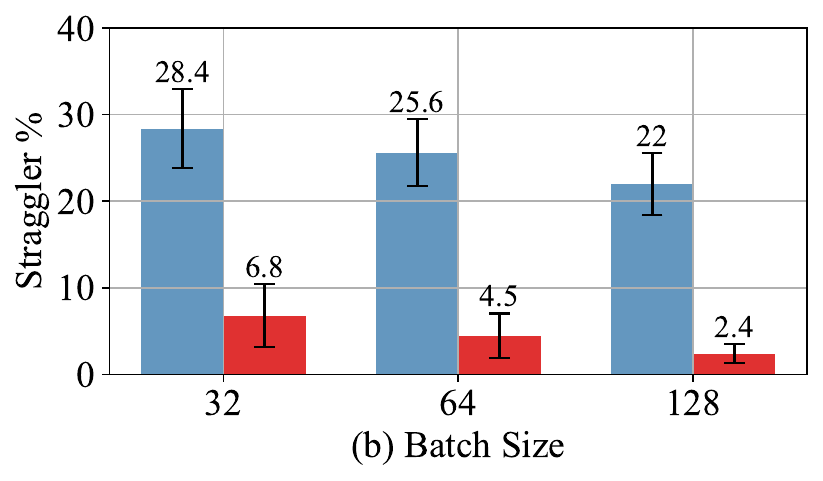}
    \end{minipage}
    \begin{minipage}[c]{0.33\textwidth}
    \centering
    \includegraphics[width=\linewidth]{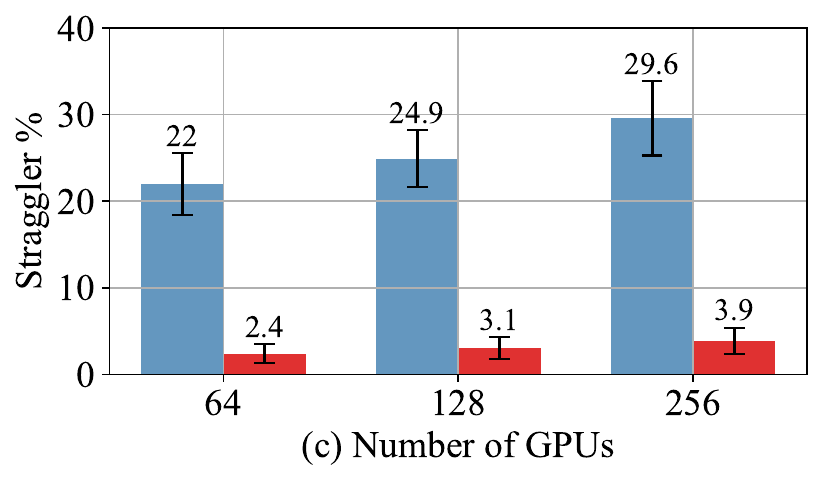}
    \end{minipage}
    \caption{Straggler Reduction}
    \label{fig:eval_straggler}
    \vspace{-1em}
\end{figure*}

For recommendation systems, the capability of fast model iteration represents a critical component for addressing evolving business requirements. 
Thus, FreeScale deliberately eschews full graph tracing, instead decomposing the training pipeline into five stages: data loading, forward propagation, backward propagation, optimizer step, and metrics computations. 
Furthermore, modern deep learning frameworks such as PyTorch provide module traversal APIs and a rich set of hook mechanisms, furnishing additional instrumentation capabilities for monitoring module performance and modifying module behaviors programmatically.
This pragmatic approach enables FreeScale to achieve a balanced compromise between optimization opportunities and development flexibility, facilitating both performance enhancement and rapid model iterations essential for industrial recommendation systems.

FreeScale identifies embedding tables via the \texttt{named\_modules()} interface, subsequently aggregating \texttt{Embedding} objects with identical \texttt{dtype} and sharding strategies into unified sharded embedding table constructs. 
The three-stage communication protocol implemented by the load balancer is inserted after sharded embedding communications, forward propagation, and backward propagation accordingly through module hooks. 
The blocking communication of the sharded embedding table is only waited in the forward function of the next iteration, which facilitates natural overlap with optimizer step, metrics computation, and data preprocessing computations. 
Consequently, the exposed communication intervals are further reduced beyond the minimal requirement of collision gradient and embedding \texttt{AllToAll} collective operations.

\section{Evaluation}\label{sec:eval}

We evaluated FreeScale using real production recommendation models and kernels running on up to 256 H100 GPUs, each with 94GB of High Bandwidth Memory (HBM). GPUs within one server are connected via $600GB/s$ NVSwitch ($4800Gb/s$) while servers are interconnected with $8 \times 200Gb/s$ InfiniBand (IB). Communications between HBM and CPU are facilitated by PCIe 5.0.

\subsection{Straggler Overhead}

Figure~\ref{fig:eval_straggler}(a) demonstrates that the straggler intensifies in TorchRec with increasing permitted UIH length. 
In contrast, FreeScale attained superior performance with longer UIH sequence, achieving more than nine-fold straggler reduction at 21,000 UIH length. 
This behavior can be attributed to the distribution of engagement history in the dataset, where the majority of users have engaged with at least 2,000 items, thus truncation to this threshold effectively eliminates workload imbalances. 
Additionally, with reduced UIH lengths, the proportion of stragglers contributed by attention kernels could be overshadowed by the jitter from other system components (e.g., data loading subsystems). 
As a result, length-based load balancing becomes less effective in short-UIH scenarios.

Figure~\ref{fig:eval_straggler}(b) illustrates that straggler percentage drops for both TorchRec and FreeScale as batch size increases, despite the corresponding increase in sparsity demonstrated in Figure~\ref{fig:background_straggler}. 
This counterintuitive result stems from the fact that the aggregate token count within a local batch constitutes the primary determinant of computation kernel execution time. 
Larger batch sizes inherently reduce cross-batch variance through statistical averaging effects, consistent with the law of large numbers. 
From an intuitive perspective, when the batch dimensions reach sufficient magnitude, all batches converge toward the intrinsic UIH length distribution of the source dataset. 
Across all evaluated configurations, FreeScale consistently outperformed TorchRec by factors ranging from four to nine.
Figure~\ref{fig:eval_straggler}(c) elucidates how scaling the computational cluster exacerbates the straggler effects. 
Straggler percentage increases for both TorchRec and FreeScale, as using more GPUs leads to higher probability of encountering outlier batches. 
All experiments presented employed FBS partitioning, because the evaluated model incorporates native variable-length UIH optimizations within custom Triton kernels, where the complexity introduced by VBS will not bring additional gains. 

\subsection{Exposed Embedding Communications}\label{eval:cosu}

Experiments in this section employ synthetic datasets for two reasons. First, to isolate the impact of prioritized communication, we exclude all other optimization techniques introduced in this paper, 
while real data suffers from stragglers.
Second, understanding the performance across varying embedding row collision ratios is essential for comprehensive evaluation. 

\begin{figure}
\begin{minipage}[c]{0.9\linewidth}
    \centering
    \includegraphics[width=\linewidth]{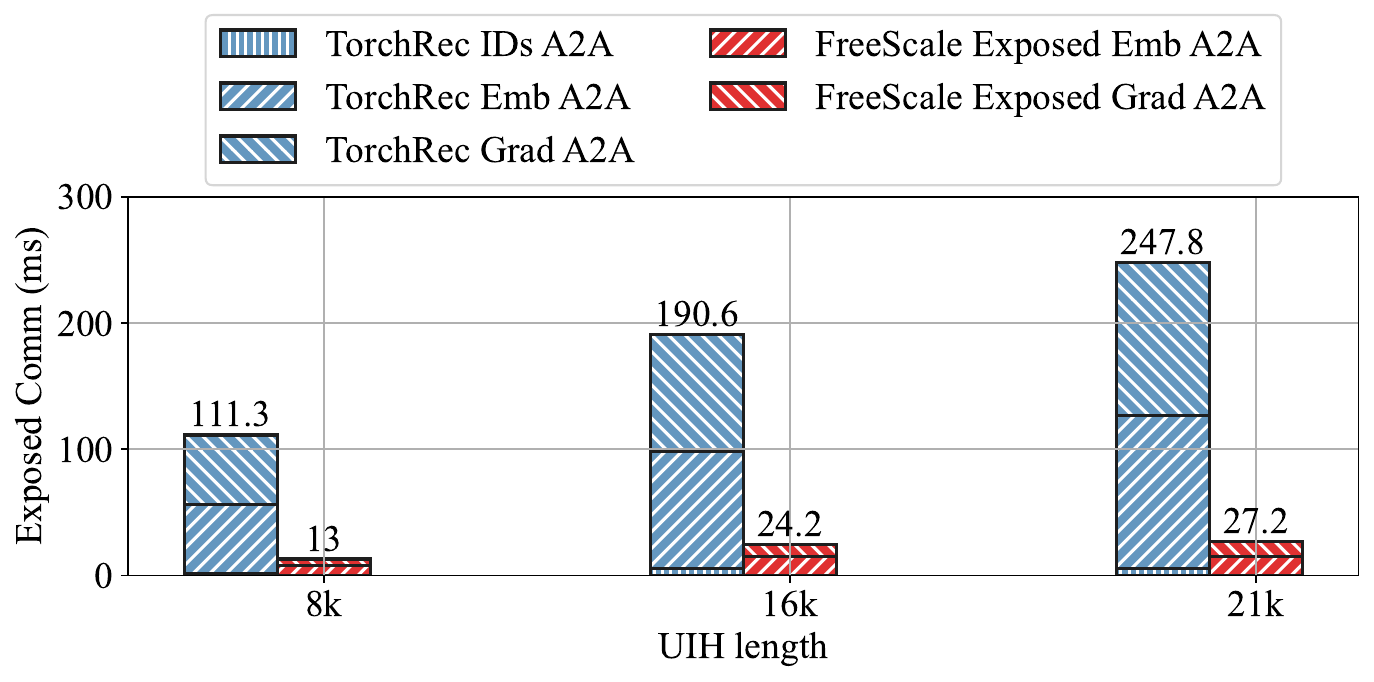}
\end{minipage}
\begin{minipage}[c]{0.9\linewidth}
    \centering
    \includegraphics[width=\linewidth]{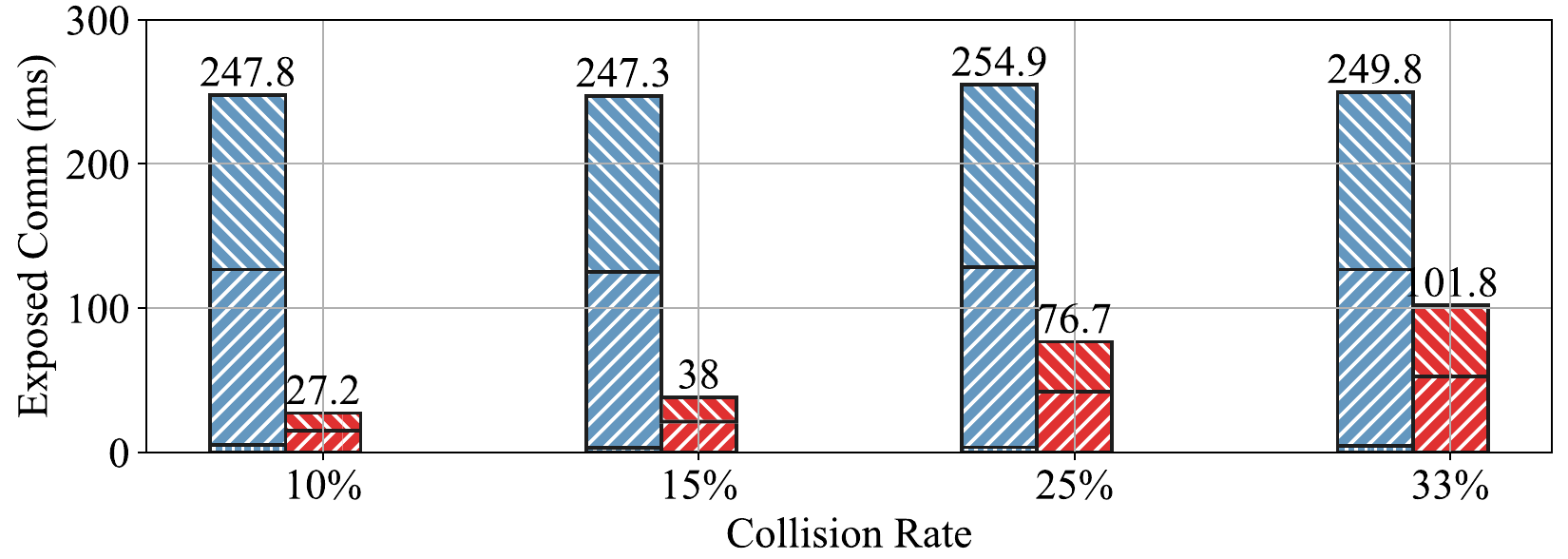}
\end{minipage}
    \caption{Isolated Exposed Communication}
    \label{fig:llm_impl}
    \vspace{-1em}
\end{figure}

As visualized in Figure~\ref{fig:llm_impl}, exposed communication is decomposed into three constituent components: IDs \texttt{AllToAll}, Embedding \texttt{AllToAll}, and Gradient \texttt{AllToAll}. 
Since FreeScale achieves complete overlap on IDs \texttt{AllToAll}  through prefetching, this component is omitted from the figure. 
When analyzing the impact of increasing UIH length, both TorchRec and FreeScale experience proportionally increased exposed communication duration, attributable to concurrent increases in both communication volume and collision ratios. 
Nevertheless, FreeScale consistently demonstrates approximately a nine-fold reduction in exposed communication latency across these configurations.
When examining performance across varying collision rates while maintaining constant communication volume, the TorchRec exhibits relatively stable exposed communication latency. 
In contrast, FreeScale demonstrates a linear relationship between exposed communication duration and collision rate, confirming that the observed latency is dominated by communicating collision rows. 
This finding aligns with the theoretical design underlying the prioritization algorithm.

\subsection{Communication and Computation Contention}

\begin{figure}[!htb]
\begin{minipage}[c]{\linewidth}
\centering
\includegraphics[width=\linewidth]{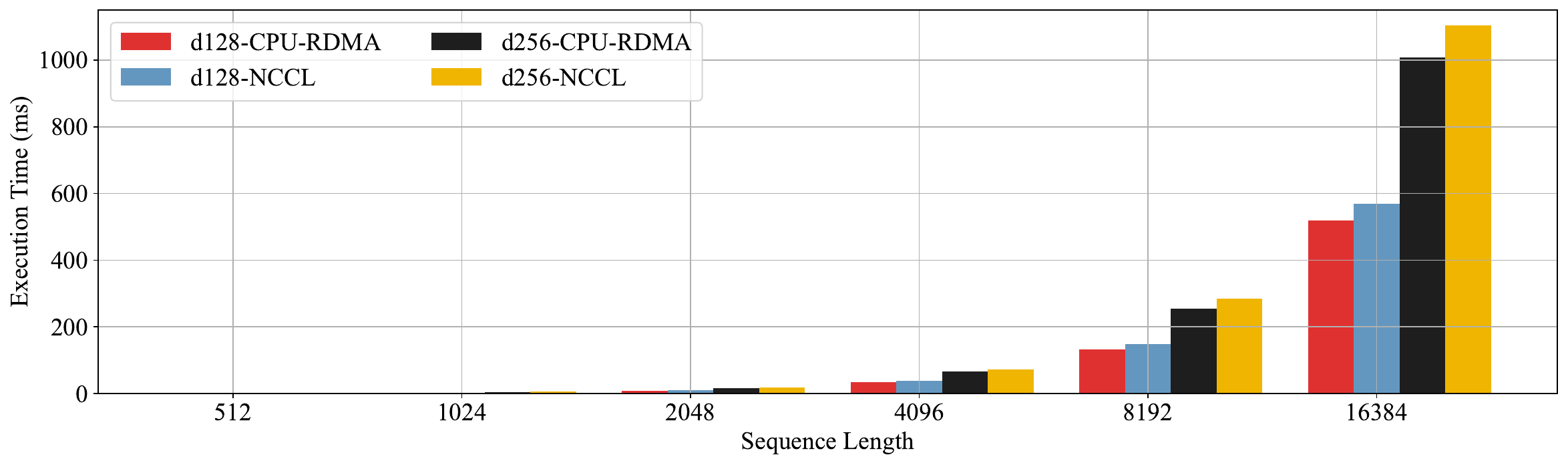}
\end{minipage}
\caption{Execution Time Comparison}
\label{fig:eval_zerosm}
\end{figure}

Figure ~\ref{fig:eval_zerosm} shows the kernel (which overlapped with communication) execution time benchmark on synthetic data.
The execution time increases exponentially as input sequence lengths increases, while growing linearly when hidden dim increases.
SM-Free communication,  outperforms NCCL communication by ~10\% across all scenarios, shows its effectiveness in mitigating communication and cumputation contention.
To be noted that, this consistent ~10\% speedup is also observed when varying the NCCL\_MAX\_NCHANNEL or NCCL\_MIN\_NCHANNEL.
We hypothesis it because NCCL's internal mechanism that automatically tune the resource utilization, while these varied settings just serve as tuning hints during runtime which saturate the hardware network bandwidth.
\subsection{Overall Speedups}\label{sec:eval_e2e}

%
%
%
%
%
%

%


\begin{figure}
\begin{minipage}[c]{\linewidth}
    \centering
    \includegraphics[width=\linewidth]{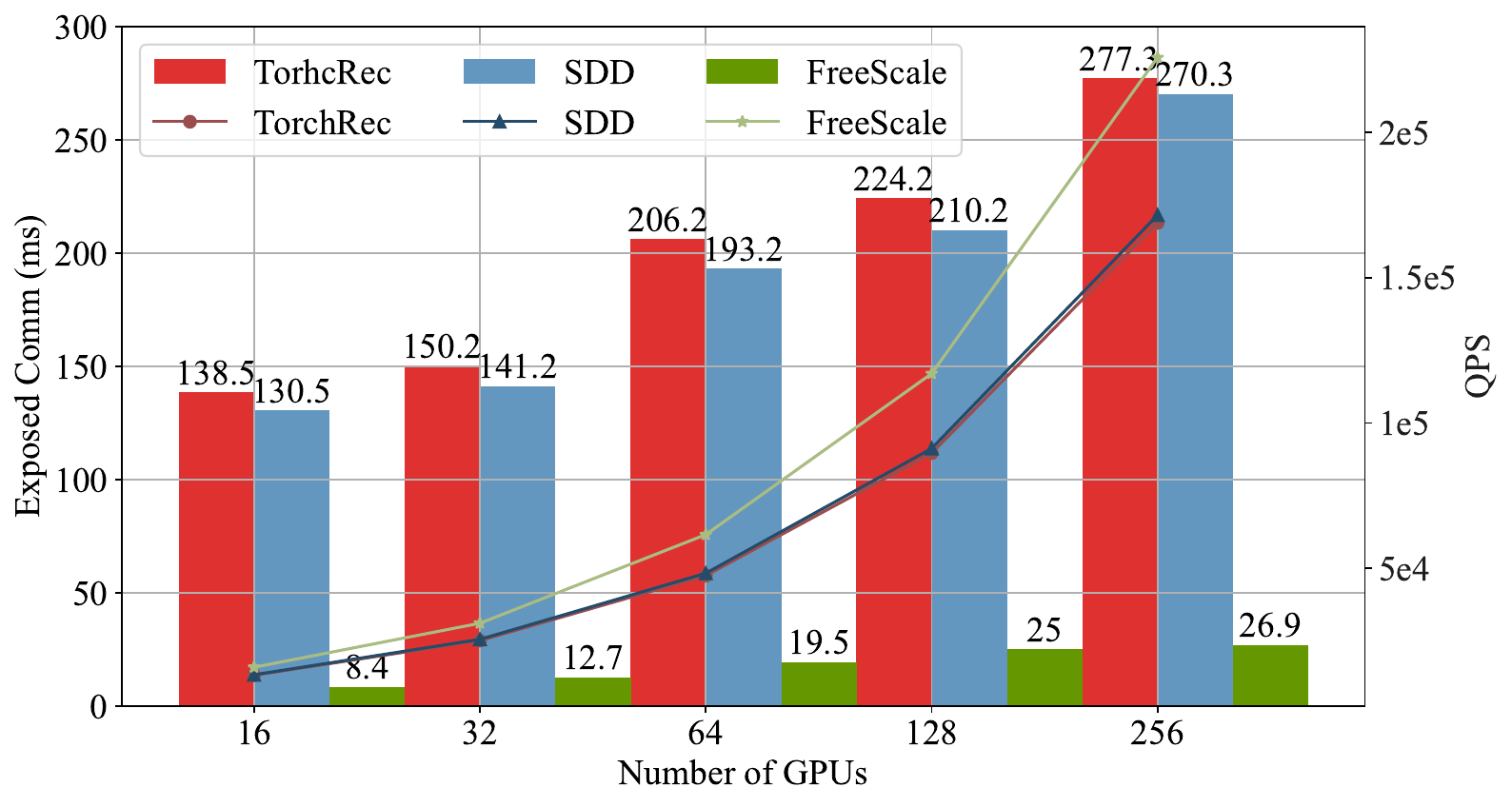}
\end{minipage}
\begin{minipage}[c]{0.95\linewidth}
    \centering
    \includegraphics[width=\linewidth]{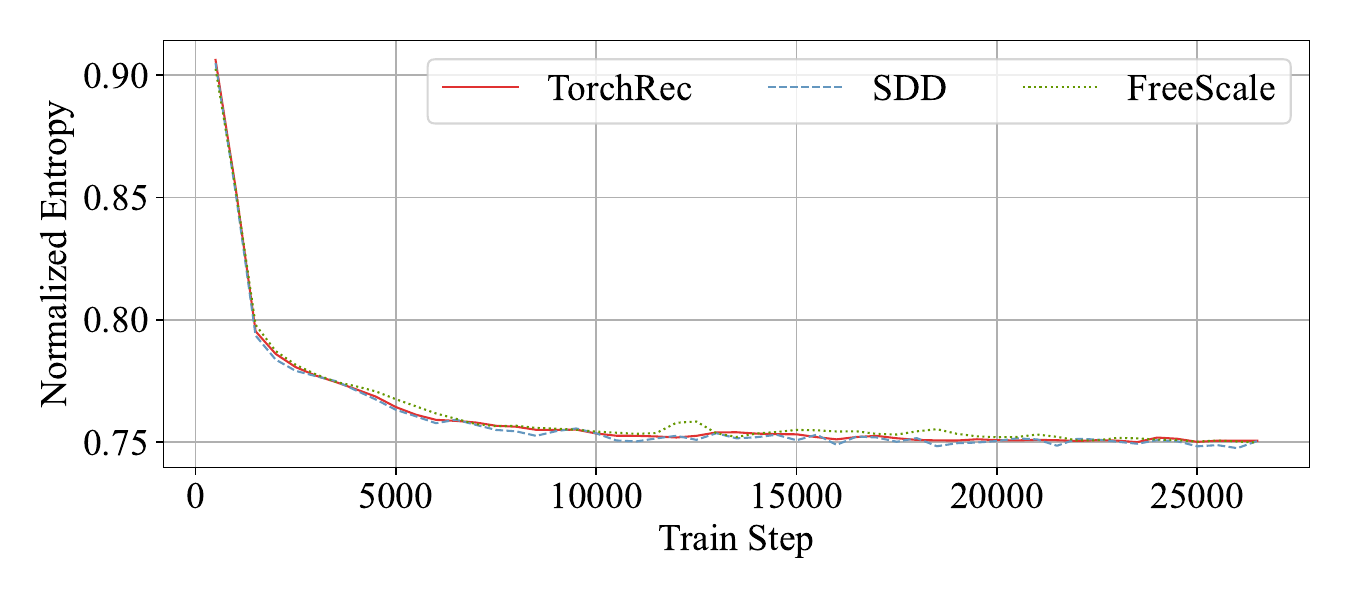}
\end{minipage}
    \caption{Overall Experiments}
    \label{fig:e2e}
    \vspace{-2em}
\end{figure}

Figure~\ref{fig:e2e} top summarizes the exposed communication latencies and corresponding throughput improvements on production data, when scaling up with larger clusters. 
The left vertical axis corresponds to the red, blue and green bars and the right vertical axis correlates with the lines. 

TorchRec, the baseline, has its exposed communication latency monotonically increases with the cluster size, 
as straggler effects intensify in larger clusters and collective communication operations suffer from reduced efficiency with larger communication topologies.

SDD, a distributed training optimization that tries to overlap distributed metadata communication with model forward operations, helps reduce some exposed communications by ~10ms. 
The communication reduction by SDD is roughly constant because the metadata communication only communicates how many elements should be transferred between GPUs and the time to transfer the number itself doesn't grow with the clusters.

In contrast, the FreeScale maintains consistently minimal exposed communication latency across cluster scales, achieving ~90\% reduction in exposed communication overhead relative to the baseline.
FreeScale's exposed communication latency also become more expensive with larger clusters, 
despite experiments maintaining constant size for both local embedding shard and UIH length. 
This can be attributed to the 
multifaceted nature of communication overhead in distributed systems. 
The total communication latency comprises three distinct components: straggler synchronization delays, network topology traversal time, and data transmission duration. 
Although FreeScale effectively controls both the straggler-induced waiting periods and data transmission volumes through advanced load balancing and prioritization techniques, 
the fundamental overhead associated with traversing increasingly complex interconnect topologies in larger clusters remains an unavoidable scaling constraint. 
%
%
%

The three curves represents the QPS (number of items processed per second) during training. TorchRec and SDD yields a roughly same QPS, because SDD only addresses metadata communication, leaving the straggler issues and embedding communication untouched. While FreeScale shows the increasing impact on larger DLRM training jobs.

Note that the experiments are conducted on clusters with high communication bandwidth ($8 \times 200Gb/s$). 
For slower connections where communication latency dominates, speedups should be more significant. 

Besides effciency improvements, an infrastructure optimization should also guarantee numeric parity with baseline. Figure~\ref{fig:e2e} bottom compares the offline training normalized entropy (NE) curve of TorchRec, SDD and FreeScale on a critical task. The results confirms that the model converges to same NE level,
confirms that FreeScale maintain numerical fidelity.
Noted that, in online training settings, FreeScale can yield more topline metric wins compared to baseline. 
It is because in online settings, model has to complete training within a predefined time window and randomly drop out unfinished training examples to avoid stuck issues. 
FreeScale enables model to learn from more training examples and thus get better performance.
\section{Discussion}\label{sec:discussion}

In this section, we will discuss the memory overhead, runtime complexity, engineering effort and the failure modes. We will also illustrate more about the SM free communication and compare to NCCL in details.
\subsection{Memory \& Time Analysis}\label{sec:discuss_memory}

As FreeScale needs to prefetch future batch to do training batch balance and pre-fetch non-collision embeddings, it introduces 1 more batches of input data.
However, since the input data only contains IDs and dense features, it usually occupies several GBs memory and much less than the memory consumption during model forward and backward computation, which usually contains 10+ transformer-like layers and each layer would create multiple tensors of similar size. So prefetching embeddings only contributes a small fraction of the overall HBM footprint.
Additionally, FreeScale also balances the memory consumption across GPUs and reduce peak memory utilization which can also make up the memory overhead introduced by itself.

In terms of time complexity, FreeScale carefully implemented a multi-GPU stream pipeline that overlaps auxiliary computation kernels with main streams, minimizing overhead through better resource utilization and therefore yield great QPS wins.

\subsection{Engineering Effort \& Failure Mode}\label{sec:discuss_engineer}
FreeScale comprises around 8,600 lines of core code  which is 2.5\% of the TorchRec codebase.
During the implementation, preserving autograd context across iterations does introduce additional engineering complexity. We wrapped that in a relatively clean way into the autograd function and our train pipeline, which makes the complexity manageable at the trainer level.

FreeScale's failure patterns (collective timeouts, OOM, stream synchronization) are identical to existing RecSys infrastructure, which already runs multi-stream GPU workloads with communication/computation overlap at scale. 
We leverage the same observability stack (Perfetto traces, memory profilers) and debugging infrastructure.
If the workload is already running at peak HBM capacity, the additional fraction of HBM footprint added by FreeScale could lead to OOM. 
However, in practice, this is usually a trade-off as HBM capacity can be restored by paying recomputation overhead through activation checkpointing. 
Another possible failure mode is where the model is using tiny embedding tables or very short sequences, then FreeScale won't bring much speedup.
\subsection{More of SM Free Communcation}\label{sec:discuss_sm_free}
In theory, when we look at communication alone without any overlapping with computations or other operations, 
NCCL-based communication should always be faster than the SM free communication as it uses more GPU resources and fewer copies. 

However, the story is different when we take a holistic view where computations are also in the picture. 
More specifically, in FreeScale, load balancing and prioritized embedding communication either introduces additional overlapped communication or transforms exposed communications into overlapped operations. 
We would like to clarify that the CPU-RDMA is proposed to do these overlapped communications without consuming SM resources. 
As NCCL- based communication needs SMs, this causes SM contention when communication overlaps with computation. 
With CPU-RDMA communication, overlapped computation runs unaffected and finishes faster.

\section{Related Work}\label{sec:related}

\subsection{Load Balancing in Distributed Training}
LB-BSP~\cite{lb_1} proposes a semi-dynamic scheme to re-balances local batch size to mitigate stragglers.
Srifty~\cite{MLSYS2022_0cafb789} uses learned compute and communication performance models to statically load balance on heterogeneous, cloud fleets.
Gandiva~\cite{222611} supports oversubscription, migration and packing to defragment cluster usage, 
improving fleet-wide utilization.
NeuroShard~\cite{neuroshard} better load balance placements of embedding tables across accelerators to minimize lookup performance.

\subsection{Optimizing DLRM Efficiency}
It's
challenging to optimize DLRM, due to the large embedding tables~\citep{lian2021persiaopenhybridscaling}, 
imbalanced workload and poorer efficiency on newer hardware~\citep{luo2018parameter, luo2020plink}.
Wukong and DHEN~\citep{zhang2024wukong, DHEN} use compressed dot products and GEMM-heavy linear compression operators for interaction mechanism.
AutoInt~\citep{song2019autoint} adopts the heavily optimized transformer architecture.
Mamba4Rec~\citep{liu2024mamba4recefficientsequentialrecommendation} uses state-space to learn sequence in linear time.
DMT~\citep{luo2024disaggregatedmultitowertopologyawaremodeling}, decomposed AlltoAll communication into a multi-rail, multi-step version and leverages faster intra-host interconnect.
PLink~\cite{luo2020plink} 
constructs hierarchical reduction trees for efficient parameter synchronization. 
Hybrid sharding schemes that leverage intra-host locality 
are also proposed in FSDP and DHEN~\cite{DHEN, zhao2023pytorchfsdpexperiencesscaling}. 
Blink~\cite{wang2018blink} constructs multiple reduction rings to optimally utilize resource.
Zero++~\cite{zero++} adopts quantized weight, gradient communication~\citep{yang2020mixed} and hierarchical weight partition. 
Specialized accelerators built for recommendation models are no longer uncommon~\cite{firoozshahian2023mtia, maddury2024next}. 
Optimal collectives synthesis~\cite{shah2025mscclrethinkinggpucommunication, 10.1145/3575693.3575724, 10.1145/3437801.3441620, 285084} creates provably optimal collective schedules from communication primitives given a target setup.
\section{Conclusion}\label{sec:conclusion}

This paper has addressed a critical efficiency challenge in distributed training of DLRMs. Our investigation identified that stragglers and blocking communications introduce substantial computational resource under-utilization which appears as computation bubbles.
FreeScale represents a holistic solution that systematically eliminates these inefficiencies through load balancing and prioritized embedding communication and resolves the GPU resource competition during the overlapping communication and computation operations via SM-Free communication techniques.
Empirical evaluation using production workloads on a cluster of 256 GPU demonstrates that FreeScale achieves a remarkable 90.3\% reduction in computational bubbles.
These results demonstrated FreeScale as an effective optimization framework for industrial-scale recommendation systems, enabling significantly improved hardware utilization without compromising model quality or numerical stability.



\bibliography{main}
\bibliographystyle{mlsys2025}

\end{document}